\documentclass{article}

\usepackage{microtype}
\usepackage{tikz, pgfplots,graphicx,xcolor}
\usetikzlibrary{plotmarks,spy,backgrounds}
\pgfplotsset{compat=newest}
\usepackage{subfigure}
\usepackage{booktabs} 
\usepackage[utf8]{inputenc}
\usepackage{amsmath,amssymb,amsthm}
\usepackage{multirow}
\usepackage{makecell}

\usepackage{hyperref}

\usepackage[accepted]{icml2018}

\newcommand{\T}{{\sf T}}
\newcommand{\E}{\mathbb{E}}

\newcommand{\X}{\mathbf{X}}
\newcommand{\C}{\mathbf{C}}
\newcommand{\W}{\mathbf{W}}
\newcommand{\G}{\mathbf{G}}
\newcommand{\U}{\mathbf{U}}
\newcommand{\B}{\mathbf{B}}
\newcommand{\V}{\mathbf{V}}
\newcommand{\A}{\mathbf{A}}
\newcommand{\M}{\mathbf{M}}
\newcommand{\J}{\mathbf{J}}
\newcommand{\Q}{\mathbf{Q}}

\newcommand{\x}{\mathbf{x}}
\newcommand{\w}{\mathbf{w}}

\newcommand{\one}{\mathbf{1}}
\newcommand{\ba}{\mathbf{a}}
\newcommand{\bb}{\mathbf{b}}

\renewcommand{\P}{\mathbf{P}}
\renewcommand{\S}{\mathbf{S}}
\renewcommand{\j}{\mathbf{j}}
\renewcommand{\t}{\mathbf{t}}

\newcommand{\bmu}{\boldsymbol{\mu}}
\newcommand{\bphi}{\boldsymbol{\phi}}
\newcommand{\bomega}{\boldsymbol{\omega}}
\newcommand{\bOmega}{\boldsymbol{\Omega}}
\newcommand{\bSigma}{\boldsymbol{\Sigma}}
\newcommand{\bPhi}{\boldsymbol{\Phi}}

\DeclareMathOperator{\tr}{tr}
\DeclareMathOperator{\ReLU}{ReLU}
\DeclareMathOperator{\LReLU}{LReLU}

\newtheorem{Assumption}{Assumption}
\newtheorem{Theorem}{Theorem}
\newtheorem{Lemma}{Lemma}
\newtheorem{Remark}{Remark}

\begin{document}

\twocolumn[
\icmltitle{On the Spectrum of Random Features Maps of High Dimensional Data}



\icmlsetsymbol{equal}{*}

\begin{icmlauthorlist}
\icmlauthor{Zhenyu Liao}{cs}
\icmlauthor{Romain Couillet}{cs,gstats}
\end{icmlauthorlist}

\icmlaffiliation{cs}{Laboratoire des Signaux et Syst{\`e}mes (L2S), CentraleSup{\'e}lec, Universit{\'e} Paris-Saclay, France;}

\icmlaffiliation{gstats}{G-STATS Data Science Chair, GIPSA-lab, University Grenobles-Alpes, France}

\icmlcorrespondingauthor{Zhenyu Liao}{zhenyu.liao@l2s.centralesupelec.fr}
\icmlcorrespondingauthor{Romain Couillet}{romain.couillet@centralesupelec.fr}

\icmlkeywords{Random Features Maps, Spectral Analysis, Random Matrix Theory, High Dimensional Statistics}

\vskip 0.3in
]



\printAffiliationsAndNotice{}  

\begin{abstract}
Random feature maps are ubiquitous in modern statistical machine learning, where they generalize random projections by means of powerful, yet often difficult to analyze nonlinear operators. In this paper, we leverage the ``concentration'' phenomenon induced by random matrix theory to perform a spectral analysis on the Gram matrix of these random feature maps, here for Gaussian mixture models of simultaneously large dimension and size. Our results are instrumental to a deeper understanding on the interplay of the nonlinearity and the statistics of the data, thereby allowing for a better tuning of random feature-based techniques.
\end{abstract}

\section{Introduction}
\label{sec:introduction}
Finding relevant features is one of the key steps for solving a machine learning problem. To this end, the backpropagation algorithm is probably the best-known method, with which superhuman performances are commonly achieved for specific tasks in applications of computer vision \cite{krizhevsky2012imagenet} and many others \cite{schmidhuber2015deep}. But data-driven approaches such as the backpropagation method, in addition to being computationally demanding, fail to cope with limited amounts of available training data.

One successful alternative in this regard is the use of ``random features'', exploited both in feed-forward neural networks \cite{huang2012extreme,scardapane2017randomness}, in large-scale kernel estimation \cite{rahimi2008random,vedaldi2012efficient} and more recently in random sketching schemes \cite{keriven2016sketching}. Random feature maps consist in projections randomly exploring the set of nonlinear representations of the data, hopefully extracting features relevant to some given task. The nonlinearities make these representations more mighty but meanwhile theoretically more difficult to analyze and optimize.

Infinitely large random features maps are nonetheless well understood as they result in (asymptotically) equivalent kernels, the most popular example being random Fourier features and their limiting radial basis kernels \cite{rahimi2008random}. Beyond those asymptotic results, recent advances in random matrix theory give rise to unexpected simplification on the understanding of the finite-dimensional version of these kernels, i.e., when the data number and size are large but of similar order as the random feature vector size \cite{el2010spectrum,couillet2016kernel}. Following the same approach, in this work, we perform a spectral analysis on the Gram matrix of the random feature matrices. This matrix is of key relevance in many associated machine learning methods (e.g., spectral clustering \cite{ng2002spectral} and kernel SVM \cite{scholkopf2002learning}) and understanding its spectrum casts an indispensable light on their asymptotic performances. In the remainder of the article, we shall constantly consider spectral clustering as a concrete example of application; however, similar analyses can be performed for other types of random feature-based algorithms.

Our contribution is twofold. From a random matrix theory perspective, it is a natural extension of the sample covariance matrix analysis \cite{silverstein1995empirical} to a nonlinear setting and can also be seen as the generalization of the recent work of \cite{pennington2017nonlinear} to a more practical data model. From a machine learning point of view, we describe quantitatively the mutual influence of different nonlinearities and data statistics on the resulting random feature maps. More concretely, based on the ratio of two coefficients from our analysis, commonly used activation functions are divided into three classes: means-oriented, covariance-oriented and balanced, which eventually allows one to choose the activation function with respect to the statistical properties of the data (or task) at hand, with a solid theoretical basis.

We show by experiments that our results, applicable theoretically only to Gaussian mixture data, show an almost perfect match when applied to some real-world datasets. We are thus optimistic that our findings, although restricted to Gaussian assumptions on the data model, can be applied to a larger set of problems beyond strongly structured ones.


\emph{Notations}: Boldface lowercase (uppercase) characters stand for vectors (matrices), and non-boldface scalars respectively. $\one_T$ is the column vector of ones of size $T$, and $\mathbf{I}_T$ the $T \times T$ identity matrix. The notation $(\cdot)^\T$ denotes the transpose operator. The norm $\| \cdot \| $ is the Euclidean norm for vectors and the operator norm for matrices.

In the remainder of this article, we introduce the objects of interest and necessary preliminaries in Section~\ref{sec:problem}. Our main results on the spectrum of random feature maps will be presented in Section~\ref{sec:main}, followed by experiments on two types of classification tasks in Section~\ref{sec:validation}. The article closes on concluding remarks and envisioned extensions in Section~\ref{sec:conclusion}.

\section{Problem Statement and Preliminaries} 
\label{sec:problem}

Let $\x_1, \ldots, \x_T \in \mathbb{R}^p$ be independent data vectors, each belonging to one of $K$ distribution classes $\mathcal{C}_1, \ldots, \mathcal{C}_K$. Class $\mathcal{C}_a$ has cardinality $T_a$, for all $a\in\{1,\ldots,K\}$. We assume that the data vector $\x_i$ follows a Gaussian mixture model\footnote{We normalize the data by $p^{-1/2}$ to guarantee that $\| \x_i \| = O(1)$ with high probability when $\| \C_a \| = O(1)$.}, i.e.,
\[
	\x_i = \bmu_a/\sqrt{p} + \bomega_i
\]
with $\bomega_i \sim \mathcal{N}(\mathbf{0},\C_a/p)$ for some mean $\bmu_a \in \mathbb{R}^{p}$ and covariance $\C_a \in \mathbb{R}^{p \times p}$ of associated class $\mathcal{C}_a$.

We denote the data matrix $\X = \begin{bmatrix} \x_1, \ldots, \x_T \end{bmatrix} \in \mathbb{R}^{p \times T}$ of size $T$ by cascading all $\x_i$ as column vectors. To extract random features, $\X$ is premultiplied by some random matrix $\W \in \mathbb{R}^{n \times p}$ with i.i.d.\@ entries and then applied entry-wise some nonlinear \emph{activation function} $\sigma(\cdot)$ to obtain the random feature matrix $\bSigma \equiv \sigma(\W\X) \in \mathbb{R}^{n \times T}$, whose columns are simply $\sigma(\W\x_i)$ the associated random feature of $\x_i$.

In this article, we focus on the Gram matrix $\G \equiv \frac1n \bSigma^\T \bSigma$ of the random features, the entry $(i,j)$ of which is given by
\[
	\G_{ij} =  \frac1n \sigma(\W \x_i)^\T \sigma(\W \x_j) = \frac1n \sum_{k = 1}^n \sigma(\w_k^\T \x_i) \sigma(\w_k^\T \x_j)
\]
with $\w_k^\T$ the $k$-th row of $\W$. Note that all $\w_k$ follow the same distribution, so that taking expectation over $\w \equiv \w_k$ of the above equation one results in the average kernel matrix $\bPhi$, with the $(i,j)$ entry of which given by
\begin{equation}
	\bPhi(\x_i, \x_j) = \E_{\w} \G_{ij}  = \E_{\w} \sigma(\w^\T \x_i) \sigma(\w^\T \x_j).
  \label{eq:definition-Phi}
\end{equation}

When the entries of $\W$ follow a standard Gaussian distribution, one can compute the generic form $\bPhi(\ba,\bb) = \E_{\w} \sigma(\w^\T \ba) \sigma(\w^\T \bb) $ by applying the integral trick from \cite{williams1997computing}, for a large set of nonlinear functions $\sigma(\cdot)$ and arbitrary vector $\ba,\bb$ of appropriate dimension. We list the results for commonly used functions in Table~\ref{tab:Phi-sigma}.

\begin{table*}[t]
\caption{$\bPhi(\ba,\bb)$ for different $\sigma(\cdot)$, $\angle(\ba,\bb) \equiv \frac{\ba^\T \bb}{\|\ba\| \|\bb\|}$.}
\label{tab:Phi-sigma}
\vskip 0.1in
\begin{center}
\begin{small}
\begin{sc}
\begin{tabular}{l|c}
\toprule
$\sigma(t)$     & $\bPhi(\ba,\bb)$ \\
\midrule
$t$    & $\ba^\T \bb$\\
\abovespace
$\max(t,0) \equiv \ReLU(t)$     & $\frac1{2\pi} \|\ba\| \|\bb\| \left(\angle(\ba,\bb)\arccos\left(-\angle(\ba,\bb)\right)+\sqrt{1-\angle(\ba,\bb)^2}\right)$   \\
\abovespace
$|t|$           & $\frac2\pi \|\ba\| \|\bb\| \left(\angle(\ba,\bb)\arcsin\left(\angle(\ba,\bb)\right)+\sqrt{1-\angle(\ba,\bb)^2}\right)$   \\
\abovespace
$\varsigma_+ \max(t,0)+ \varsigma_- \max(-t,0)$ & $ \frac12 (\varsigma_+^2 + \varsigma_-^2) \ba^\T\bb + \frac{\|\ba\| \|\bb\|}{2\pi} (\varsigma_+ + \varsigma_-)^2 \left( \sqrt{1 - \angle(\ba,\bb)^2 } - \angle(\ba,\bb) \arccos(\angle(\ba,\bb)) \right) $\\
\abovespace
$1_{t>0}$       & $\frac12 -\frac1{2\pi} \arccos\left(\angle(\ba,\bb)\right)$    \\
\abovespace
${\rm sign}(t)$ & $\frac2\pi\arcsin\left(\angle(\ba,\bb)\right)$    \\
\abovespace
$\varsigma_2 t^2 + \varsigma_1 t + \varsigma_0$ & $\varsigma_2^2 \left( 2\left( \ba^\T \bb \right)^2 + \|\ba\|^2 \|\bb\|^2 \right) + \varsigma_1^2 \ba^\T \bb + \varsigma_2 \varsigma_0 \left( \|\ba\|^2 + \|\bb\|^2 \right) + \varsigma_0^2$\\
\abovespace
$\cos(t)$       & $\exp\left(-\frac12 \left(\|\ba\|^2 + \|\bb\|^2\right)\right) \cosh(\ba^\T \bb)$  \\
\abovespace
$\sin(t)$       & $\exp\left(-\frac12 \left(\|\ba\|^2 + \|\bb\|^2\right)\right) \sinh(\ba^\T \bb)$  \\
\abovespace
${\rm erf}(t)$  & $\frac2\pi \arcsin\left(\frac{2\ba^\T \bb}{\sqrt{(1+2\|\ba\|^2)(1+2\|\bb\|^2)}}\right)$    \\
\belowspace
$\exp(-\frac{t^2}2)$ & $\frac1{ \sqrt{ (1+\|\ba\|^2) (1+\|\bb\|^2) - (\ba^\T \bb)^2 } }$\\
\bottomrule
\end{tabular}
\end{sc}
\end{small}
\end{center}
\vskip -0.1in
\end{table*}

Since the Gram matrix $\G$ describes the correlation of data in the \emph{feature space}, it is natural to recenter $\G$, and thus $\bPhi$ by pre- and post-multiplying a projection matrix $\P \equiv \mathbf{I}_T - \frac1T \one_T \one_T^\T$. In the case of $\bPhi$, we get
\[
	\bPhi_c \equiv \P \bPhi \P.
\]
In the recent line of works \cite{louart2018random,pennington2017nonlinear}, it has been shown that the large dimensional (large $n,p,T$) characterization of $\G$, in particular its eigenspectrum, is fully determined by $\bPhi$ and the ratio $n/p$. For instance, by defining the \emph{empirical spectral distribution} of $\G_c = \P \G \P$ as $\rho_{\G_c}(x) \equiv \frac1T \sum_{i=1}^T 1_{\lambda_i \le x} (x)$, with $\lambda_1, \ldots, \lambda_T$ the eigenvalues of $\G_c$, it has been shown in \cite{louart2018random} that, as $n,p,T \to \infty$, $\rho_{\G}(x)$ almost surely converges to a non-random distribution $\rho(x)$, referred to as the \emph{limiting spectral distribution} of $\G_c$ such that
\[
  \rho(x) = \frac1{\pi} \lim_{y\to 0^+} \int_{-\infty}^x \Im \left[ m(t + i y) \right] dt.
\] 
with $m(z)$ the associated Stieltjes transform given by
\[
  m(z) = \frac1n \tr \Q(z), \ \Q(z) \equiv \left( \frac{\bPhi_c}{1+\delta(z)} - z \mathbf{I}_T \right)^{-1} 
\]
with $\delta(z)$ the unique solution of $\delta(z) = \frac1n \tr \left(\bPhi_c \Q(z) \right)$.


As a consequence, in the objective of understanding the asymptotic behavior of $\G_c$ as $n,p,T$ are simultaneously large, we shall focus our analysis on $\bPhi_c$. To this end, the following assumptions will be needed throughout the paper.
\begin{Assumption}[Growth rate]
    As $T \to \infty$, 
    \begin{description}
        \item[$1)$] $p/T \to c_0 \in (0,\infty)$,
        \item[$2)$] for each $a \in \{1,\ldots,K\}$, $T_a/T \to c_a \in (0,1)$,
        \item[$3)$] $\| \bmu_a \| = O(1)$,
        \item[$4)$] let $\C^\circ \equiv \sum_{a=1}^K \frac{T_a}{T} \C_a$ and for $a \in \{1,\ldots,K\}$, $\C_a^\circ \equiv \C_a - \C^\circ$, then $\| \C_a\| = O(1)$ and $\tr(\C_a^\circ)/\sqrt{p} = O(1)$,
        \item[$5)$] for technical convenience we assume in addition that $\tau \equiv \tr\left(\C^\circ \right)/p $ converges in $(0, \infty)$.
    \end{description}
    \label{ass:Growth-rate}
\end{Assumption}

Assumption~\ref{ass:Growth-rate} ensures that the information of means or covariances is neither too simple nor impossible to be extracted from the data, as investigated in \cite{couillet2016kernel}.

Let us now introduce the key steps of our analysis. Under Assumption~\ref{ass:Growth-rate}, note that for $\x_i \in \mathcal{C}_a$ and $\x_j \in \mathcal{C}_b$, $i \neq j$, 
\[
  \x_i^\T \x_j = \underbrace{ \bomega_i^\T \bomega_j}_{O(p^{-1/2})} + \underbrace{ \bmu_a^\T \bmu_b /p + \bmu_a^\T \bomega_j/\sqrt{p} + \bmu_b^\T \bomega_i / \sqrt{p}}_{O(p^{-1})}
\]
which allows one to perform a Taylor expansion around $0$ as $p,T \to \infty$, to give a reasonable approximation of nonlinear functions of $\x_i^\T \x_j$, such as those appearing in $\bPhi_{ij}$ (see again Table~\ref{tab:Phi-sigma}). For $i=j$, one has instead
\[
  \| \x_i\|^2 =  \underbrace{ \|\bomega_i\|^2 }_{O(1)} + \underbrace{ \| \bmu_a\|^2/p +  2 \bmu_a^\T \bomega_i/ \sqrt{p} }_{O(p^{-1})}.
\]

From $\E_{\bomega_i} [\| \bomega_i\|^2] = \tr(\C_a)/p$ it is convenient to further write $\| \bomega_i\|^2 = \tr(\C_a) /p + \left( \| \omega_i\|^2 - \tr(\C_a)/p \right)$, where $ \tr(\C_a) /p = O(1)$ and $\| \omega_i\|^2 - \tr(\C_a) /p = O(p^{-1/2})$. By definition $\tau \equiv \tr(\C^\circ)/p = O(1)$ and exploiting again Assumption~\ref{ass:Growth-rate} one results in,
\begin{align*}
  \|\x_i\|^2 &= \underbrace{\tau}_{O(1)} + \underbrace{ \tr (\C_a^\circ) /p + \|\bomega_i\|^2 - \tr (\C_a) / p}_{O(p^{-1/2})} \\
  &+ \underbrace{ \| \bmu_a\|^2/p + 2\bmu_a^\T \bomega_i /\sqrt{p}}_{O(p^{-1})}
\end{align*}
which allows for a Taylor expansion of nonlinear functions of $\|\x_i\|^2$ around $\tau$, as has been done for $\x_i^\T \x_j$.

From Table~\ref{tab:Phi-sigma}, it appears that, for every listed $\sigma(\cdot)$, $\bPhi(\x_i,\x_j)$ is a smooth function of $\x_i^\T \x_j$ and $\| \x_i \|$, $\| \x_j \|$, despite their possible discontinuities (e.g., the $\ReLU$ function and $\sigma(t) = |t|$). The above results thus allow for an entry-wise Taylor expansion of the matrix $\bPhi$ in the large $p,T$ limit.

A critical aspect of the analysis where random matrix theory comes into play now consists in developing $\bPhi$ as a sum of matrices arising from the Taylor expansion and ignoring terms that give rise to a vanishing operator norm, so as to find an asymptotic equivalent matrix $\tilde \bPhi$ such that $\| \bPhi - \tilde \bPhi \| \to 0 $ as $p,T \to \infty$, as described in detail in the following section. This analysis provides a simplified asymptotically equivalent expression for $\bPhi$ with all nonlinearities removed, which is the crux of the present study.

\section{Main Results}
\label{sec:main}


In the remainder of this article, we shall use the following notations for random elements,
\[
	\bOmega \equiv \begin{bmatrix} \bomega_1, \ldots, \bomega_T \end{bmatrix}, \bphi \equiv \left\{ \| \bomega_i\|^2 - \E \| \bomega_i \|^2 \right\}_{i=1}^T
\]
such that $\bOmega \in \mathbb{R}^{p \times T}$,  $\bphi \in \mathbb{R}^T$. For deterministic elements\footnote{As a reminder here, $\M$ stands for \emph{means}, $\t$ accounts for (difference in) \emph{traces} while $\S$ for the ``\emph{shapes}'' of covariances.},
\begin{align*}
\M &\equiv \begin{bmatrix} \bmu_1, \ldots, \bmu_K \end{bmatrix} \in \mathbb{R}^{p \times K}, \t \equiv \left\{ \tr \C_a^\circ/\sqrt{p} \right\}_{a=1}^K \\
\J &\equiv \begin{bmatrix} \j_1, \ldots, \j_K \end{bmatrix}\in \mathbb{R}^{T \times K}, \S \equiv \left\{ \tr(\C_a \C_b) /p \right\}_{a,b=1}^K 
\end{align*}
with $\t \in \mathbb{R}^K$, $\S \in \mathbb{R}^{K \times K}$ and $\j_a\in \mathbb{R}^T$ denotes the canonical vector of class $\mathcal{C}_a$ such that $(\j_a)_i = \delta_{\x_i \in \mathcal{C}_a}$.

\begin{Theorem}[Asymptotic Equivalent of $\bPhi_c$] Let Assumption~\ref{ass:Growth-rate} hold and $\bPhi_c$ be defined as $\bPhi_c \equiv \P \bPhi \P$, with $\bPhi$ given in \eqref{eq:definition-Phi}. Then, as $T \to \infty$, for all $\sigma(\cdot)$ given in Table~\ref{tab:Phi-sigma},\footnote{For all functions $\sigma(\cdot)$ listed in Table~\ref{tab:Phi-sigma} we identified a ``pattern'' in the structure of $\tilde{\bPhi}_c$, which then led to Theorem~\ref{theo:asymp-equiv-of-Phi-c} and Table~\ref{tab:coef-Phi-c}. This two-step approach does not yet allow to justify whether this pattern goes beyond these (listed) functions; hence Theorem~\ref{theo:asymp-equiv-of-Phi-c} is stated so far solely for these functions.}
\[
  \|\bPhi_c - \tilde \bPhi_c \| \to 0
\]
almost surely, with $\tilde \bPhi_c = \P \tilde \bPhi \P$ and
\[
  \tilde \bPhi \equiv d_1 \left( \bOmega + \M \frac{\J^\T}{\sqrt{p}}  \right)^\T \left(\bOmega + \M \frac{\J^\T}{\sqrt{p}} \right) + d_2 \U \B \U^\T + d_0 \mathbf{I}_T
\]
where we recall that $\P \equiv \mathbf{I}_T - \frac1T \one_T \one_T^\T$ and
\[
	\U \equiv \begin{bmatrix} \frac{\J}{\sqrt{p}}, \bphi \end{bmatrix},\quad \B \equiv \begin{bmatrix} \t\t^\T + 2\S & \t \\ \t^\T & 1 \end{bmatrix}
\]

with the coefficients $d_0, d_1, d_2$ given in Table~\ref{tab:coef-Phi-c}.
\label{theo:asymp-equiv-of-Phi-c}
\end{Theorem}
We refer the readers to Section~\ref{sm:proof-main-theo} in Supplementary Material for a detailed proof of Theorem~\ref{theo:asymp-equiv-of-Phi-c}.

\begin{table*}[t]
\caption{Coefficients $d_i$ in $\tilde \bPhi_c$ for different $\sigma(\cdot)$.}
\label{tab:coef-Phi-c}
\vskip 0.1in
\begin{center}
\begin{small}
\begin{sc}
\begin{tabular}{l|c|c|c}
\toprule
$\sigma(t)$ & $d_0$ & $d_1$ & $d_2$ \\
\midrule
$t$   & 0 & 1 & 0 \\
\abovespace
$\max(t,0)\ \equiv \ReLU(t)$  & $\left(\frac14 - \frac1{2\pi} \right)\tau$ & $\frac14$ & $\frac1{8\pi \tau}$  \\
\abovespace
$|t|$  & $\left(1 - \frac2\pi \right)\tau$ & 0 & $\frac1{2\pi \tau}$ \\
\abovespace
$\varsigma_+ \max(t,0) + \varsigma_- \max(-t,0)\ \equiv \LReLU(t)$ & $\frac{\pi - 2}{4\pi} (\varsigma_+ + \varsigma_-)^2 \tau$ & $\frac14 (\varsigma_+ - \varsigma_-)^2$ & $\frac1{8\tau\pi}(\varsigma_+ + \varsigma_-)^2$ \\
\abovespace
$1_{t>0}$  & $\frac14 - \frac1{2\pi}$ & $\frac1{2\pi\tau}$ & 0  \\
${\rm sign}(t)$  & $1 - \frac2\pi$ & $\frac2{\pi\tau}$ & 0 \\
\abovespace
$\varsigma_2 t^2 + \varsigma_1 t + \varsigma_0$ & $2\tau^2 \varsigma_2^2$ & $\varsigma_1^2$ & $\varsigma_2^2$ \\
\abovespace
$\cos(t)$        & $\frac12 + \frac{e^{-2\tau}}2 - e^{-\tau}$ & 0 & $\frac{e^{-\tau}}4$ \\
\abovespace
$\sin(t)$        & $\frac12 - \frac{e^{-2\tau}}2 - \tau e^{-\tau}$ & $e^{-\tau}$ & 0 \\
\abovespace
${\rm erf}(t)$  & $\frac2\pi \left( \arccos\left( \frac{2\tau}{2\tau+1} \right) - \frac{2\tau}{2\tau+1} \right)$ & $\frac4\pi \frac1{2\tau + 1}$ & 0 \\
\abovespace
\belowspace
$\exp(-\frac{t^2}2)$       & $\frac1{\sqrt{2\tau+1}} - \frac1{\tau+1}$ & 0 & $\frac1{4(\tau+1)^3}$ \\
\bottomrule
\end{tabular}
\end{sc}
\end{small}
\end{center}
\vskip -0.1in
\end{table*}

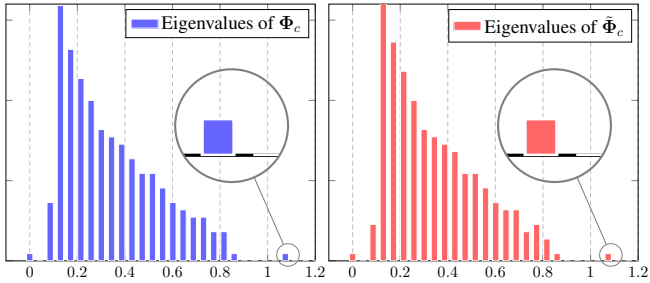
\begin{figure}[htb]
\vskip 0.1in
\begin{center}
\begin{minipage}[b]{0.48\columnwidth}%
\begin{tikzpicture}[scale=0.6,spy using outlines]
\renewcommand{\axisdefaulttryminticks}{4} 
\pgfplotsset{every major grid/.style={densely dashed}}       
\tikzstyle{every axis y label}+=[yshift=-10pt] 
\tikzstyle{every axis x label}+=[yshift=5pt]
\pgfplotsset{every axis legend/.append style={cells={anchor=west},fill=white, at={(0.98,0.98)}, anchor=north east, font=\large }}
\begin{axis}[
xmin=-.1,
ymin=0,
xmax=1.2,
ymax=3.2,
bar width=4pt,
grid=major,
ymajorgrids=false,
scaled ticks=true,
yticklabels = {},
]
\addplot+[ybar,mark=none,color=white,fill=blue!60!white,area legend] coordinates{
(0.000000,0.090914)(0.042966,0.000000)(0.085933,0.727313)(0.128899,3.181996)(0.171865,2.636511)(0.214832,2.272855)(0.257798,2.000112)(0.300764,1.636455)(0.343731,1.545541)(0.386697,1.454627)(0.429663,1.272799)(0.472630,1.090970)(0.515596,1.090970)(0.558562,0.909142)(0.601529,0.727313)(0.644495,0.636399)(0.687462,0.545485)(0.730428,0.545485)(0.773394,0.363657)(0.816361,0.363657)(0.859327,0.090914)(0.902293,0.000000)(0.945260,0.000000)(0.988226,0.000000)(1.031192,0.000000)(1.074159,0.090914)(1.117125,0.000000)(1.160091,0.000000)(1.203058,0.000000)(1.246024,0.000000)(1.288990,0.000000)(1.331957,0.000000)(1.374923,0.000000)(1.417889,0.000000)(1.460856,0.000000)(1.503822,0.000000)(1.546788,0.000000)(1.589755,0.000000)(1.632721,0.000000)(1.675687,0.000000)(1.718654,0.000000)(1.761620,0.000000)(1.804586,0.000000)(1.847553,0.000000)(1.890519,0.000000)(1.933486,0.000000)(1.976452,0.000000)(2.019418,0.000000)(2.062385,0.000000)(2.105351,0.000000)
};
\addlegendentry{{Eigenvalues of $\bPhi_c$}}
\begin{scope}
    \spy[black!50!white,size=1.5cm,circle,connect spies,magnification=5] on (3.75,0.08) in node [fill=none] at (5,3);
\end{scope}
\end{axis}
\end{tikzpicture}
\end{minipage}
\hfill{}
\begin{minipage}[b]{0.48\columnwidth}
\begin{tikzpicture}[scale=0.6,spy using outlines]
\renewcommand{\axisdefaulttryminticks}{4} 
\pgfplotsset{every major grid/.style={densely dashed}}       
\tikzstyle{every axis y label}+=[yshift=-10pt] 
\tikzstyle{every axis x label}+=[yshift=5pt]
\pgfplotsset{every axis legend/.append style={cells={anchor=west},fill=white, at={(0.98,0.98)}, anchor=north east, font=\large }}
\begin{axis}[
xmin=-0.1,
ymin=0,
xmax=1.2,
ymax=3.2,
yticklabels = {},
bar width=4pt,
grid=major,
ymajorgrids=false,
scaled ticks=true,
]
\addplot+[ybar,mark=none,color=white,fill=red!60!white,area legend] coordinates{
(0.000000,0.090914)(0.042966,0.000000)(0.085933,0.454571)(0.128899,3.272910)(0.171865,2.727425)(0.214832,2.363769)(0.257798,2.000112)(0.300764,1.636455)(0.343731,1.545541)(0.386697,1.454627)(0.429663,1.363713)(0.472630,1.090970)(0.515596,1.090970)(0.558562,0.909142)(0.601529,0.727313)(0.644495,0.636399)(0.687462,0.636399)(0.730428,0.363657)(0.773394,0.454571)(0.816361,0.272743)(0.859327,0.090914)(0.902293,0.000000)(0.945260,0.000000)(0.988226,0.000000)(1.031192,0.000000)(1.074159,0.090914)(1.117125,0.000000)(1.160091,0.000000)(1.203058,0.000000)(1.246024,0.000000)(1.288990,0.000000)(1.331957,0.000000)(1.374923,0.000000)(1.417889,0.000000)(1.460856,0.000000)(1.503822,0.000000)(1.546788,0.000000)(1.589755,0.000000)(1.632721,0.000000)(1.675687,0.000000)(1.718654,0.000000)(1.761620,0.000000)(1.804586,0.000000)(1.847553,0.000000)(1.890519,0.000000)(1.933486,0.000000)(1.976452,0.000000)(2.019418,0.000000)(2.062385,0.000000)(2.105351,0.000000)
};
\addlegendentry{{Eigenvalues of $\tilde\bPhi_c$}}
\begin{scope}
    \spy[black!50!white,size=1.5cm,circle,connect spies,magnification=5] on (3.75,0.08) in node [fill=none] at (5,3);
\end{scope}
\end{axis}
\end{tikzpicture}
\end{minipage}
\caption{Eigenvalue distribution of $\bPhi_c$ and $\tilde \bPhi_c$ for the $\ReLU$ function and Gaussian mixture data  with $\bmu_a = \begin{bmatrix}\mathbf{0}_{a-1};3;\mathbf{0}_{p-a}\end{bmatrix}$, $\C_a= \left(1 + 2(a-1)/\sqrt{p} \right)\mathbf{I}_p$, $p=512$, $T = 256$ and $c_1 = c_2 = 1/2$. Expectation estimated with $500$ realizations of $\W$.}
\label{fig:eigvenvalue-Gc-tGc-ReLU}
\end{center}
\vskip -0.1in
\end{figure}

\begin{figure}[htb]
\vskip 0.1in
\begin{center}
\begin{tikzpicture}[spy using outlines]
\renewcommand{\axisdefaulttryminticks}{4} 
\tikzstyle{every axis y label}+=[yshift=-10pt] 
\tikzstyle{every axis x label}+=[yshift=5pt]
\pgfplotsset{every axis legend/.append style={cells={anchor=east},fill=white, at={(0.02,0.98)}, anchor=north west, font=\scriptsize }}
\begin{axis}[
width=\columnwidth,
height=.6\columnwidth,
xmin = 0,
xmax = 128,
ymax = 0.2,
ytick={-0.1,-0.05,0,0.05,0.1},
yticklabels = {$-0.1$,$-0.05$,$0$,$0.05$,$0.1$},
xticklabels = {},
bar width=4pt,
scaled ticks=true,
]
\addplot[color=blue!60!white,line width=0.5pt] coordinates{
(1.000000, -0.067304)(2.000000, -0.018178)(3.000000, -0.047135)(4.000000, -0.021948)(5.000000, -0.048043)(6.000000, -0.041147)(7.000000, -0.062194)(8.000000, -0.032799)(9.000000, -0.030261)(10.000000, -0.062559)(11.000000, -0.071643)(12.000000, -0.034116)(13.000000, -0.028880)(14.000000, -0.065518)(15.000000, -0.023358)(16.000000, -0.012985)(17.000000, -0.063115)(18.000000, -0.037394)(19.000000, -0.103634)(20.000000, -0.030804)(21.000000, -0.071456)(22.000000, -0.057920)(23.000000, -0.060385)(24.000000, -0.021771)(25.000000, -0.057825)(26.000000, -0.108134)(27.000000, 0.005264)(28.000000, -0.007370)(29.000000, -0.064567)(30.000000, -0.066989)(31.000000, -0.032220)(32.000000, -0.076447)(33.000000, -0.019075)(34.000000, -0.096344)(35.000000, -0.046206)(36.000000, -0.004870)(37.000000, -0.054868)(38.000000, -0.088634)(39.000000, -0.080658)(40.000000, -0.053694)(41.000000, -0.046904)(42.000000, -0.044128)(43.000000, -0.101142)(44.000000, -0.012559)(45.000000, -0.066500)(46.000000, -0.060567)(47.000000, -0.052505)(48.000000, -0.075528)(49.000000, -0.064402)(50.000000, -0.106125)(51.000000, 0.010510)(52.000000, -0.068981)(53.000000, -0.071756)(54.000000, -0.046392)(55.000000, 0.003052)(56.000000, -0.067969)(57.000000, -0.011722)(58.000000, -0.023997)(59.000000, 0.002702)(60.000000, -0.082942)(61.000000, -0.131870)(62.000000, -0.041320)(63.000000, -0.042652)(64.000000, -0.117633)(65.000000, 0.079292)(66.000000, 0.005214)(67.000000, 0.062738)(68.000000, 0.074142)(69.000000, -0.027830)(70.000000, 0.087120)(71.000000, 0.055942)(72.000000, -0.001250)(73.000000, 0.082111)(74.000000, 0.095583)(75.000000, 0.015278)(76.000000, 0.065347)(77.000000, 0.062737)(78.000000, 0.047942)(79.000000, 0.080769)(80.000000, 0.067515)(81.000000, 0.057637)(82.000000, -0.009677)(83.000000, 0.066538)(84.000000, -0.009200)(85.000000, 0.002145)(86.000000, 0.053373)(87.000000, 0.099783)(88.000000, 0.060830)(89.000000, 0.056095)(90.000000, 0.035448)(91.000000, 0.109505)(92.000000, -0.017733)(93.000000, 0.030046)(94.000000, 0.112193)(95.000000, 0.052479)(96.000000, 0.099625)(97.000000, 0.075368)(98.000000, 0.046731)(99.000000, 0.088096)(100.000000, 0.048083)(101.000000, 0.034669)(102.000000, 0.051871)(103.000000, 0.050175)(104.000000, 0.092156)(105.000000, 0.053239)(106.000000, 0.133993)(107.000000, -0.009298)(108.000000, 0.011092)(109.000000, 0.024430)(110.000000, 0.005511)(111.000000, 0.057133)(112.000000, 0.104488)(113.000000, 0.068635)(114.000000, 0.073827)(115.000000, 0.021323)(116.000000, 0.011160)(117.000000, 0.059617)(118.000000, 0.054423)(119.000000, 0.043766)(120.000000, -0.009641)(121.000000, 0.087035)(122.000000, 0.011492)(123.000000, 0.067501)(124.000000, 0.053244)(125.000000, 0.000394)(126.000000, 0.074765)(127.000000, 0.055727)(128.000000, 0.007250)
};
\addlegendentry{{ Leading eigenvector of $\bPhi_c$ }}
\addplot[densely dashed,color=red!60!white,line width=1pt] coordinates{
(1.000000, -0.065054)(2.000000, -0.016096)(3.000000, -0.047720)(4.000000, -0.023370)(5.000000, -0.055497)(6.000000, -0.042125)(7.000000, -0.049214)(8.000000, -0.034383)(9.000000, -0.023770)(10.000000, -0.056379)(11.000000, -0.079389)(12.000000, -0.040094)(13.000000, -0.028908)(14.000000, -0.075024)(15.000000, -0.018285)(16.000000, -0.009405)(17.000000, -0.049953)(18.000000, -0.036522)(19.000000, -0.103097)(20.000000, -0.027478)(21.000000, -0.078960)(22.000000, -0.062605)(23.000000, -0.061972)(24.000000, -0.034115)(25.000000, -0.037815)(26.000000, -0.103893)(27.000000, -0.002701)(28.000000, -0.019161)(29.000000, -0.064876)(30.000000, -0.064173)(31.000000, -0.020725)(32.000000, -0.081927)(33.000000, -0.022171)(34.000000, -0.100842)(35.000000, -0.039325)(36.000000, -0.011046)(37.000000, -0.067076)(38.000000, -0.082311)(39.000000, -0.081459)(40.000000, -0.052305)(41.000000, -0.044561)(42.000000, -0.054468)(43.000000, -0.108393)(44.000000, -0.007697)(45.000000, -0.070246)(46.000000, -0.050934)(47.000000, -0.044608)(48.000000, -0.068999)(49.000000, -0.069141)(50.000000, -0.091803)(51.000000, 0.006479)(52.000000, -0.071482)(53.000000, -0.069112)(54.000000, -0.050921)(55.000000, -0.006555)(56.000000, -0.052171)(57.000000, -0.011324)(58.000000, -0.025418)(59.000000, -0.011501)(60.000000, -0.074884)(61.000000, -0.131433)(62.000000, -0.039806)(63.000000, -0.037632)(64.000000, -0.120890)(65.000000, 0.077883)(66.000000, 0.003433)(67.000000, 0.070010)(68.000000, 0.084754)(69.000000, -0.030441)(70.000000, 0.078783)(71.000000, 0.058191)(72.000000, -0.001653)(73.000000, 0.079065)(74.000000, 0.095938)(75.000000, 0.012139)(76.000000, 0.051804)(77.000000, 0.075443)(78.000000, 0.053225)(79.000000, 0.090003)(80.000000, 0.069797)(81.000000, 0.050533)(82.000000, -0.008457)(83.000000, 0.055785)(84.000000, -0.003415)(85.000000, 0.001459)(86.000000, 0.038279)(87.000000, 0.097325)(88.000000, 0.073952)(89.000000, 0.053573)(90.000000, 0.038011)(91.000000, 0.107705)(92.000000, -0.011189)(93.000000, 0.047543)(94.000000, 0.109944)(95.000000, 0.049572)(96.000000, 0.083397)(97.000000, 0.065893)(98.000000, 0.047363)(99.000000, 0.083135)(100.000000, 0.052571)(101.000000, 0.034728)(102.000000, 0.033688)(103.000000, 0.052767)(104.000000, 0.078222)(105.000000, 0.032748)(106.000000, 0.138068)(107.000000, -0.006874)(108.000000, 0.021021)(109.000000, 0.020421)(110.000000, 0.000913)(111.000000, 0.059977)(112.000000, 0.104708)(113.000000, 0.069036)(114.000000, 0.074540)(115.000000, 0.014593)(116.000000, 0.013782)(117.000000, 0.057370)(118.000000, 0.065781)(119.000000, 0.073520)(120.000000, -0.023298)(121.000000, 0.094394)(122.000000, 0.005392)(123.000000, 0.062296)(124.000000, 0.052416)(125.000000, 0.006074)(126.000000, 0.085046)(127.000000, 0.054346)(128.000000, -0.001982)
};
\addlegendentry{{ Leading eigenvector of $\tilde \bPhi_c$ }}
\begin{scope}
    \spy[black!50!white,size=1.2cm,circle,connect spies,magnification=3] on (6.25,2.1) in node [fill=none] at (5.8,0.85);
\end{scope}
\end{axis}
\node (A) at (2,0.5) {$\mathcal{C}_1$};
\node (B) at (5,0.5) {$\mathcal{C}_2$};
\draw[<->] (0.02,0.2) -- (3.4,0.2);
\draw[<->] (3.5,0.2) -- (6.62,0.2);
\end{tikzpicture}
\caption{Leading eigenvector of $\bPhi_c$ and $\tilde \bPhi_c$ in the settings of Figure~\ref{fig:eigvenvalue-Gc-tGc-ReLU}, with $\j_1 = \begin{bmatrix} \mathbf{1}_{T_1}; \mathbf{0}_{T_2} \end{bmatrix}$ and $\j_2 = \begin{bmatrix} \mathbf{0}_{T_1}; \mathbf{1}_{T_2} \end{bmatrix}$.}
\label{fig:eigenvector-Gc-tGc-ReLU}
\end{center}
\vskip -0.1in
\end{figure}
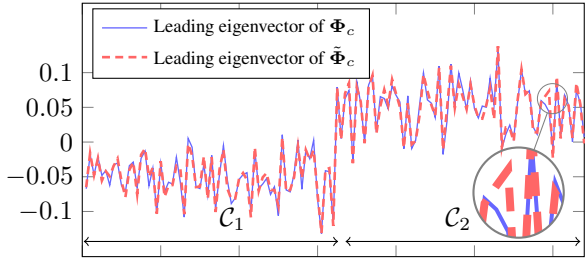 

Theorem~\ref{theo:asymp-equiv-of-Phi-c} tells us as a corollary (see Corollary~4.3.15 in \cite{horn2012matrix}) that the maximal difference between the eigenvalues of $\bPhi_c$ and $\tilde \bPhi_c$ vanishes asymptotically as $p,T \to \infty$, as confirmed in Figure~\ref{fig:eigvenvalue-Gc-tGc-ReLU}. Similarly the distance between the ``isolated eigenvectors\footnote{Eigenvectors that correspond to the eigenvalues found at a non-vanishing distance from the other eigenvalues.}'' also vanishes, as seen in Figure~\ref{fig:eigenvector-Gc-tGc-ReLU}. This is of tremendous importance as the determination of the leading eigenvalues and eigenvectors of $\bPhi_c$ (that contain crucial information for clustering, for example) can be studied from the equivalent problem performed on $\tilde \bPhi_c$ and becomes mathematically more tractable.

On closer inspection of Theorem~\ref{theo:asymp-equiv-of-Phi-c}, the matrix $\tilde \bPhi$ is expressed as the sum of three terms, weighted respectively by the three coefficients $d_0, d_1$ and $d_2$, that depend on the nonlinear function $\sigma(\cdot)$ via Table~\ref{tab:coef-Phi-c}. Note that the statistical structure of the data $\{\x_i\}_{i=1}^T$ (namely the means in $\M$ and the covariances in $\t$ and $\S$) is perturbed by random fluctuations ($\bOmega$ and $\bphi$) and it is thus impossible to get rid of these noisy terms by wisely choosing the function $\sigma(\cdot)$. This is in sharp contrast to \cite{couillet2016kernel} where it is shown that more general kernels (i.e., not arising from random feature maps) allow for a more flexible treatment of information versus noise.

\begin{figure}[htb]
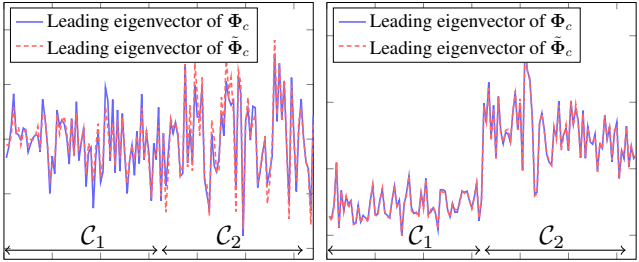

\vskip 0.1in
\begin{center}
\begin{minipage}[!t]{0.48\columnwidth}%

\end{minipage}
\caption{Leading eigenvector of $\bPhi_c$ and $\tilde \bPhi_c$ for ${\rm erf}$ (left) and the $\ReLU$ (right) function, performed on Gaussian mixture data  with $\bmu_a = \mathbf{0}_p$, $\C_a= \left(1 + 15(a-1)/\sqrt{p}\right)\mathbf{I}_p$, $p=512$, $T = 256$, $c_1 = c_2 = 1/2$; $\j_1 = %
$. Expectation estimated by averaging over $500$ realizations of $\W$.}
\label{fig:eigenvector-Gc-tGc-erf}
\end{center}
\vskip -0.1in
\end{figure}

However, there does exist a balance between the means and covariances, that provides some instructions in the appropriate choice of the nonlinearity. From Table~\ref{tab:coef-Phi-c}, the functions $\sigma(\cdot)$ can be divided into the following three groups:
\begin{itemize}
  \item \emph{mean-oriented}, where $d_1 \neq 0$ while $d_2 = 0$: this is the case of the functions $t$, $1_{t>0}$, ${\rm sign}(t)$, $\sin(t)$ and ${\rm erf}(t)$, which asymptotically track only the difference in means (i.e., $\t$ and $\S$ disappear from the expression of $\tilde\bPhi_c$). As an illustration, in Figure~\ref{fig:eigenvector-Gc-tGc-erf} one fails to separate two Gaussian datasets of common mean but of different covariances with the ${\rm erf}$ function, while $\ReLU$ is able to accomplish the task;
  \item \emph{covariance-oriented}, where $d_1 = 0$ while $d_2 \neq 0$: this concerns the functions $|t|$, $\cos(t)$ and $\exp ( -t^2/2)$, which asymptotically track only the difference in covariances. Figure~\ref{fig:eigvenvalue-and-vector-Gc-tGc-abs} illustrates the impossibility to classify Gaussian mixture with same covariance with $\sigma(t) = |t|$, in contrast to the $\ReLU$ function;
  \item \emph{balanced}, where both $d_1,d_2 \neq 0$: here for the $\ReLU$ function $\max(t,0)$, the Leaky $\ReLU$ function \cite{maas2013rectifier} $\varsigma_+ \max(t,0) + \varsigma_- \max(-t,0)$ and the quadratic function $\varsigma_2 t^2 + \varsigma_1 t + \varsigma_0$.
\end{itemize}

%
%
%
%
%
%
%
%
\begin{figure}[htb]
\vskip 0.1in
\begin{center}
\begin{minipage}[!t]{0.48\columnwidth}%
%
%
\end{minipage}
\caption{Leading eigenvector of $\bPhi_c$ and $\tilde \bPhi_c$ for $\sigma(t) = |t|$ (left) and the $\ReLU$ (right) function, performed on Gaussian mixture data with $\bmu_a = %
$. Expectation estimated by averaging over $500$ realizations of $\W$.}
\label{fig:eigvenvalue-and-vector-Gc-tGc-abs}
\end{center}
\vskip -0.1in
\end{figure}

Before entering into a more detailed discussion of Theorem~\ref{theo:asymp-equiv-of-Phi-c}, first note importantly that, for practical interests, the quantity $\tau$ can be estimated consistently from the data, as described in the following lemma.
\begin{Lemma}[Consistent estimator of $\tau$]
Let Assumption~\ref{ass:Growth-rate} hold and recall the definition $\tau \equiv \tr \left( \C^\circ \right)/p$. Then, as $T \to \infty$, with probability $1$
\[
	\frac1T \sum_{i=1}^T \| \x_i \|^2 - \tau \to 0.
\]
\label{lem:consistent-estimator-of-tau}
\end{Lemma}
\vspace{-1cm}
\begin{proof}
Since
\[
	\frac1T \sum_{i=1}^T \| \x_i \|^2 = \frac1T \sum_{a=1}^K \sum_{i=1}^{T_a} \frac1p \| \bmu_a \|^2 - \frac2{\sqrt{p}} \bmu_a^\T \bomega_i + \| \bomega_i \|^2,
\]
with Assumption~\ref{ass:Growth-rate} we have $\frac1T \sum_{a=1}^K \sum_{i=1}^{T_a} \frac1p \| \bmu_a \|^2 = O(p^{-1})$. The term $ \frac1T \sum_{a=1}^K \sum_{i=1}^{T_a} \frac2{\sqrt{p}} \bmu_a^\T \bomega_i$ is a linear combination of independent zero-mean Gaussian variables and vanishes with probability $1$ as $p,T \to \infty$ with Chebyshev's inequality and the Borel-Cantelli lemma. Ultimately by the strong law of large numbers, we have $\frac1T \sum_{i=1}^{T} \| \bomega_i \|^2 - \tau \to 0$ almost surely, which concludes the proof.
\end{proof}

From a practical aspect, a few remarks on the conclusions of Theorem~\ref{theo:asymp-equiv-of-Phi-c} can be made.

\begin{Remark}[Constant shift in feature space]
For $\sigma(t) = \varsigma_2 t^2 + \varsigma_1 t + \varsigma_0$, note the absence of $\varsigma_0$ in Table~2, meaning that the constant of the quadratic function does not affect the spectrum of the feature matrix. More generally, it can be shown through the integral trick of \cite{williams1997computing} that the function $\sigma(t) + c$ for some constant shift $c$ gives the same matrix $\bPhi_c$ as the original function $\sigma(t)$.
\label{rem:constant-shift}
\end{Remark}

A direct consequence of Remark~\ref{rem:constant-shift} is that the coefficients $d_0, d_1, d_2$ of the function ${\rm sign}(t)$ are four times those of $1_{t>0}$, as a result of the fact that ${\rm sign}(t) = 2 \cdot 1_{t>0} - 1$. Constant shifts have, as such, no consequence in classification or clustering applications.

\begin{Remark}[Universality of quadratic and Leaky $\ReLU$ functions]
Ignoring the coefficient $d_0$ that gives rise to a constant shift of all eigenvalues of $\tilde \bPhi_c$ and thus of no practical relevance, observe from Table~\ref{tab:coef-Phi-c} that by tuning the parameters of the quadratic and Leaky $\ReLU$ functions ($\LReLU(t)$), one can select arbitrary positive value for the ratio $d_1/d_2$, while the other listed functions have constraints linking $d_1$ to $d_2$.
\label{rem:universality}
\end{Remark}

Following the discussions in Remark~\ref{rem:universality}, the parameters $\varsigma_+, \varsigma_-$ of the $\LReLU$, as well as $\varsigma_1, \varsigma_2$ of the quadratic function, essentially act to balance the weights of means and covariances in the mixture model of the data. More precisely, as $\frac{\varsigma_+}{\varsigma_-} \to 1$ or $\varsigma_2\gg\varsigma_1$, more emphasis is set on the ``distance'' between covariance matrices while $\frac{\varsigma_+}{\varsigma_-} \to -1$ or $\varsigma_1\gg\varsigma_2$ stresses the differences in means.



\begin{figure}[htb]
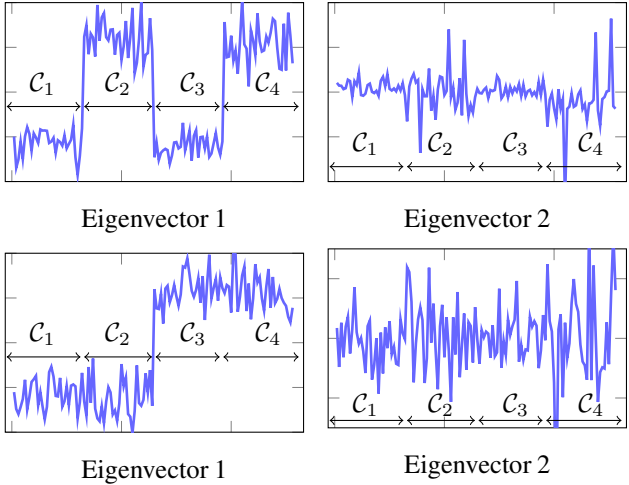

\vskip 0.1in
\begin{center}
\begin{minipage}[!t]{0.48\columnwidth}%
\centering
%
\end{minipage}
\caption{Leading two eigenvectors of $\bPhi_c$ for the $\LReLU$ function with $\varsigma_+ = -\varsigma_- = 1$ (top) and $\varsigma_+ = \varsigma_- = 1$ (bottom), performed on four classes Gaussian mixture data with $p=512$, $T = 256$, $c_a = \frac14$ and $\j_a = %
$, for $a=1,2,3,4$. Expectation estimated by averaging over $500$ realizations of $\W$.}
\label{fig:clustering-with-LReLU}
\end{center}
\vskip -0.1in
\end{figure}
\begin{figure}[h!]
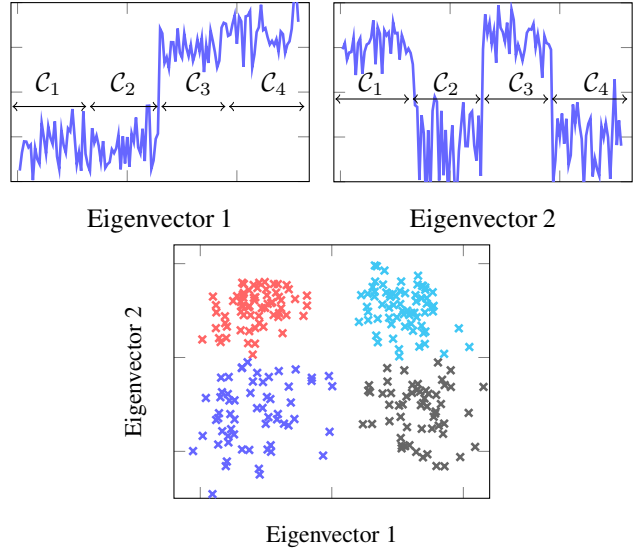

\vskip 0.1in
\begin{center}
\begin{minipage}[!t]{0.48\columnwidth}%
\centering
%
\caption{Leading two eigenvectors of $\bPhi_c$ (top) for the $\LReLU$ function with $\varsigma_+ = 1 $, $ \varsigma_- = 0$ and two dimensional representation of these eigenvectors (bottom), in the same setting as in Figure~\ref{fig:clustering-with-LReLU}.}
\label{fig:clustering-with-ReLU}
\end{center}
\vskip -0.1in
\end{figure}

In Figure~\ref{fig:clustering-with-LReLU}, spectral clustering on four classes of Gaussian data is performed: $\mathcal{N}(\bmu_1, \C_1)$, $\mathcal{N}(\bmu_1, \C_2)$, $\mathcal{N}(\bmu_2, \C_1)$ and $\mathcal{N}(\bmu_2, \C_2)$ with the $\LReLU$ function that takes different values for $\varsigma_+$ and $\varsigma_-$. For $a=1,2$, $\bmu_a = \begin{bmatrix}\mathbf{0}_{a-1};5;\mathbf{0}_{p-a}\end{bmatrix}$ and $\C_a= \left(1 + \frac{15(a-1)}{\sqrt{p}}\right)\mathbf{I}_p$. By choosing $\varsigma_+ = -\varsigma_- = 1$ (equivalent to $\sigma(t) = |t|$) and $\varsigma_+ = \varsigma_- = 1$ (equivalent to the linear map $\sigma(t) = t$), with the leading two eigenvectors we always recover two classes instead of four, as each setting of parameters only allows for a part of the statistical information of the data to be used for clustering. However, by taking $\varsigma_+ = 1,\varsigma_- = 0$ (the $\ReLU$ function) we distinguish all four classes in the leading two eigenvectors, to which the k-means method can then be applied for final classification, as shown in Figure~\ref{fig:clustering-with-ReLU}.

%
%
%
%

%
%
%
%

Of utmost importance for random feature-based spectral methods (such as kernel spectral clustering discussed above \cite{ng2002spectral}) is the presence of informative eigenvectors in the spectrum of $\G$, and thus of $\bPhi_c$. To gain a deeper understanding on the spectrum of $\bPhi_c$, one can rewrite $\tilde \bPhi$ in the more compact form,
\begin{equation}\label{eq:spike-model}
	\tilde \bPhi = d_1 \bOmega^\T \bOmega + \V \A \V^\T + d_0 \mathbf{I}_T
\end{equation}
where
\[
	\V \equiv \begin{bmatrix} \frac{\J}{\sqrt{p}}, \bphi, \bOmega^\T \M \end{bmatrix}, \ \A \equiv \begin{bmatrix} \A_{11} & d_2\t & d_1 \mathbf{I}_K \\ d_2\t^\T & d_2 & 0 \\ d_1 \mathbf{I}_K & 0 & 0 \end{bmatrix}
\]
with \( \A_{11} \equiv d_1 \M^\T \M + d_2(\t\t^\T + 2\S) \), that is akin to the so-called ``spiked model'' in the random matrix literature \cite{baik2005phase}, as it equals, if $d_1 \neq 0$, the sum of some standard (noise-like) random matrix $\bOmega^\T \bOmega$, and a low rank (here up to $2K+1$) informative matrix $\V\A\V^\T$, that may induce some isolated eigenvalues outside the main bulk of eigenvalues in the spectrum of $\tilde \bPhi_c$, as shown in Figure~\ref{fig:eigvenvalue-Gc-tGc-ReLU}.

The eigenvectors associated to these eigenvalues often contain crucial information about the data statistics (the classes in a classification settings). In particular, note that the matrix $\V$ contains the canonical vector $\j_a$ of class $\mathcal{C}_a$ and we thus hope to find some isolated eigenvector of $\bPhi_c$ aligned to $\j_a$ that can be directly used to perform clustering. Intuitively speaking, if the matrix $\A$ contains sufficient energy (has sufficiently large operator norm), the eigenvalues associated to the small rank matrix $\V\A\V^\T$ may jump out from the main bulk of $\bOmega^\T \bOmega$ and becomes ``isolated'' as in Figure~\ref{fig:eigvenvalue-Gc-tGc-ReLU}, referred to as the \emph{phase transition} phenomenon in the random matrix literature \cite{baik2005phase}. The associated eigenvectors then tend to align to linear combinations of the canonical vectors $\j_a$ as seen in Figure~\ref{fig:clustering-with-LReLU}-\ref{fig:clustering-with-ReLU}. This alignment between the isolated eigenvectors and $\j_a$ is essentially measured by the amplitude of the eigenvalues of the matrix $\A_{11}$, or more concretely, the statistical differences of the data (namely, $\t$, $\S$ and $\M$). Therefore, a good adaptation of the ratio $d_1/d_2$ ensures the (asymptotic) detectability of different classes from the spectrum of $\bPhi_c$.

\section{Numerical Validations}
\label{sec:validation}

We complete this article by showing that our theoretical results, derived from Gaussian mixture models, show an unexpected close match in practice when applied to some real-world datasets. We consider two different types of classification tasks: one on handwritten digits of the popular MNIST \cite{lecun1998mnist} database (number $6$ and $8$), and the other on epileptic EEG time series data \cite{andrzejak2001indications} (set B and E). These two datasets are typical examples of means-dominant (handwritten digits recognition) and covariances-dominant (EEG times series classification) tasks. This is numerically confirmed in Table~\ref{tab:estimation-statistics-data}. Python 3 codes to reproduce the results in this section are available at \url{https://github.com/Zhenyu-LIAO/RMT4RFM}.

\begin{table}[h!]
\caption{Empirical estimation of (normalized) differences in
means and covariances of the MNIST (Figure~\ref{fig:eigenvalue-MNIST} and~\ref{fig:eigenvector-MNIST}) and epileptic EEG (Figure~\ref{fig:eigenvalue-EEG} and~\ref{fig:eigenvector-EEG}) datasets.}
\label{tab:estimation-statistics-data}
\vskip 0.11in
\begin{center}
\begin{small}
\begin{sc}
\begin{tabular}{lcc}
\toprule
 & $\| \M^\T \M \|$ & $ \| \t\t^\T + 2\S \|$ \\
\midrule
MNIST data & $172.4$  & $86.0$  \\ 
EEG data & $1.2$  & $182.7$ \\ 
\bottomrule
\end{tabular}
\end{sc}
\end{small}
\end{center}
\vskip -0.1in
\end{table}

\subsection{Handwritten digits recognition}

\begin{figure}[htb]
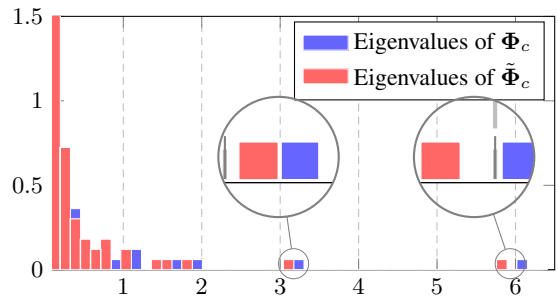

\vskip 0.1in
\begin{center}

\caption{Eigenvalue distribution of $\bPhi_c$ and $\tilde \bPhi_c$ for the MNIST data, with the $\ReLU$ function, $p=784$, $T=128$ and $c_1 = c_2 = \frac12$, with $\j_1 = %
$. Expectation estimated by averaging over $500$ realizations of $\W$. }
\label{fig:eigenvalue-MNIST}
\end{center}
\vskip -0.1in
\end{figure}

We perform random feature-based spectral clustering on data matrices that consist of $T=32,64$ and $128$ randomly selected vectorized images of size $p = 784$ from the MNIST dataset. The ``true'' means and covariances are empirically obtained from the full set of $11\,769$ MNIST images ($5\,918$ images of number $6$ and $5\,851$ of number $8$) to construct the matrix $\tilde \bPhi_c$ as per Theorem~\ref{theo:asymp-equiv-of-Phi-c}. Comparing the matrix $\bPhi_c$ built from the data and the theoretically equivalent $\tilde \bPhi_c$ obtained as if the data were Gaussian with the (empirically) computed means and covariances, we observe an extremely close fit in the behavior of the eigenvalues in Figure~\ref{fig:eigenvalue-MNIST}, as well of the leading eigenvector in Figure~\ref{fig:eigenvector-MNIST}. The k-means method is then applied to the leading two eigenvectors of the matrix $\G_c \in \mathbb{R}^{T \times T}$ that consists of $n=32$ random features to perform unsupervised classification, with resulting accuracies (averaged over $50$ runs) reported in Table~\ref{tab:MNIST-results}. As remarked from Table~\ref{tab:estimation-statistics-data}, the mean-oriented $\sigma(t)$ functions are expected to outperform the covariance-oriented ones in this task, which is consistent with the results in Table~\ref{tab:MNIST-results}.
%
%
%
%
%
%
\begin{figure}[htb]
\vskip 0.1in
\begin{center}
%
\caption{Leading eigenvector of $\bPhi_c$ for the MNIST and Gaussian mixture data with a width of $\pm 1$ standard deviations (generated from $500$ trials) in the settings of Figure~\ref{fig:eigenvalue-MNIST}.}
\label{fig:eigenvector-MNIST}
\end{center}
\vskip -0.1in
\end{figure} 

\begin{table}[t]
\caption{Classification accuracies for random feature-based spectral clustering with different $\sigma(t)$ on the MNIST dataset.}
\label{tab:MNIST-results}
\vskip 0.1in
\begin{center}
\begin{small}
\begin{sc}
\begin{tabular}{l|cccc}
\toprule
& $\sigma(t)$ & $T=32$ & $T=64$ & $T=128$ \\
\midrule
\multirow{5}{*}{\makecell{mean-\\oriented}} & $t$ & $85.31\%$  & $\mathbf{88.94\%}$  & $87.30\%$ \\
& $1_{t>0}$ & $86.00\%$  & $82.94\%$  & $85.56\%$ \\
& ${\rm sign}(t)$ & $81.94\%$  & $83.34\%$  & $85.22\%$ \\
& $\sin(t)$ & $85.31\%$  & $87.81\%$  & $\mathbf{87.50\%}$ \\ 
& ${\rm erf}(t)$ & $\mathbf{86.50\%}$  & $87.28\%$  & $86.59\%$ \\ 
\midrule
\multirow{3}{*}{\makecell{cov-\\oriented}} & $|t|$ & $62.81\%$  & $60.41\%$  & $57.81\%$ \\
& $\cos(t)$ & $62.50\%$  & $59.56\%$  & $57.72\%$ \\
& $\exp(-\frac{t^2}2)$ & $64.00\%$ & $60.44\%$ & $58.67\%$ \\
\midrule
balanced & $\ReLU(t)$ & $82.87\%$  & $85.72\%$  & $82.27\%$ \\ 
\bottomrule
\end{tabular}
\end{sc}
\end{small}
\end{center}
\vskip -0.1in
\end{table}

\subsection{EEG time series classification}

\begin{figure}[htb]
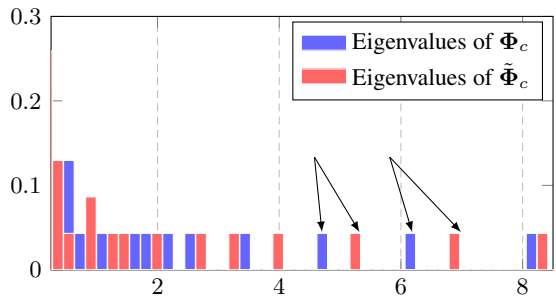

\vskip 0.1in
\begin{center}

\caption{Eigenvalue distribution of $\bPhi_c$ and $\tilde \bPhi_c$ for the epileptic EEG, with the $\ReLU$ function, $p=100$, $T=128$ and $c_1 = c_2 = \frac12$, with $\j_1 = %
$. Expectation estimated by averaging over $500$ realizations of $\W$. }
\label{fig:eigenvalue-EEG}
\end{center}
\vskip -0.2in
\end{figure}

The epileptic EEG dataset\footnote{\url{http://www.meb.unibonn.de/epileptologie/science/physik/eegdata.html}.}, developed by the University of Bonn, Germany, is described in \cite{andrzejak2001indications}. The dataset consists of five subsets (denoted A-E), each containing 100 single-channel EEG segments of 23.6-sec duration. Sets A and B were collected from surface EEG recordings of five healthy volunteers, while sets C, D and E were collected from the EEG records of the pre-surgical diagnosis of five epileptic patients. Here we perform random feature-based spectral clustering on $T=32,64$ and $128$ randomly picked EEG segments of length $p=100$ from the dataset. Means and covariances are empirically estimated from the full set ($4\,097$ segments of set B and $4\,097$ segments of set E). Similar behavior of eigenpairs as for Gaussian mixture models is once more observed in Figure~\ref{fig:eigenvalue-EEG} and~\ref{fig:eigenvector-EEG}. After k-means classification on the leading two eigenvectors of the (centered) Gram matrix composed of $n=32$ random features, the accuracies (averaged over $50$ runs) are reported in Table~\ref{tab:EEG-results}.

As opposed to the MNIST image recognition task, from Table~\ref{tab:EEG-results} it is easy to check that the covariance-oriented functions (i.e., $\sigma(t) = |t|$, $\cos(t)$ and $\exp(-t^2/2)$) far outperform any other with almost perfect classification accuracies. It is particularly interesting to note that the popular $\ReLU$ function is suboptimal in both tasks, but never performs very badly, thereby offering a good risk-performance tradeoff.

%
%
%
%
%
%
\begin{figure}[htb]
\vskip 0.2in
\begin{center}
%
\caption{Leading eigenvector of $\bPhi_c$ for the EEG and Gaussian mixture data with a width of $\pm 1$ standard deviation (generated from $500$ trials) in the settings of Figure~\ref{fig:eigenvalue-EEG}.}
\label{fig:eigenvector-EEG}
\end{center}
\vskip -0.1in
\end{figure}

\begin{table}[h!]
\caption{Classification accuracies for random feature-based spectral clustering with different $\sigma(t)$ on the epileptic EEG dataset.}
\label{tab:EEG-results}
\vskip 0.1in
\begin{center}
\begin{small}
\begin{sc}
\begin{tabular}{l|cccc}
\toprule
& $\sigma(t)$ & $T=32$ & $T=64$ & $T=128$ \\
\midrule
\multirow{5}{*}{\makecell{mean-\\oriented}} & $t$ & $71.81\%$  & $70.31\%$  & $69.58\%$ \\ 
& $1_{t>0}$ & $65.19\%$  & $65.87\%$  & $63.47\%$ \\
& ${\rm sign}(t)$ & $67.13\%$  & $64.63\%$  & $63.03\%$ \\
& $\sin(t)$ & $71.94\%$  & $70.34\%$  & $68.22\%$ \\ 
& ${\rm erf}(t)$ & $69.44\%$  & $70.59\%$  & $67.70\%$ \\
\midrule
\multirow{3}{*}{\makecell{cov-\\oriented}} & $|t|$ & $99.69\%$  & $99.69\%$  & $99.50\%$ \\
& $\cos(t)$ & $99.00\%$  & $99.38\%$  & $99.36\%$ \\
& $\exp(-\frac{t^2}2)$ & $\mathbf{99.81\%}$  & $\mathbf{99.81\%}$  & $\mathbf{99.77\%}$ \\
\midrule
balanced & $\ReLU(t)$ & $84.50\%$  & $87.91\%$  & $90.97\%$ \\ 
\bottomrule
\end{tabular}
\end{sc}
\end{small}
\end{center}
\vskip -0.1in
\end{table}

\section{Conclusion}
\label{sec:conclusion}

In this article, we have provided a theoretical analysis on random feature-based spectral algorithms for large dimensional data, providing a better understanding of the precise mechanism underlying these methods. Our results show a quite simple relation between the nonlinear function involved in the random feature map (only through two scalars $d_1$ and $d_2$) and the capacity of the latter to discriminate data upon their means and covariances. In obtaining this result, we demonstrated that point-wise nonlinearities can be incorporated into a classical Taylor expansion as a consequence of the concentration phenomenon in high dimensional space. This result was then validated through experimental classification tasks on the MNIST and EEG datasets.

Although Theorem~\ref{theo:asymp-equiv-of-Phi-c} is stated here solely for the functions listed in Table~\ref{tab:Phi-sigma}, our current line of investigation consists in directly linking the activation function $\sigma(\cdot)$ and the coefficients $d_0$, $d_1$ and $d_2$ in Table~\ref{tab:coef-Phi-c} so as to generalize our results and to provide more insights into the attributes of a function that makes it mean- or covariance-oriented; this undertaking is however more technically demanding but still likely achievable through the extension of existing results related to the work of \cite{cheng2013spectrum}. 

From a point of view of clustering, the crucial information to distinguish different classes is contained in the isolated eigenvalue/eigenvector pairs as shown in \eqref{eq:spike-model}, the asymptotic behavior of these pairs, as well as their significance for clustering are technically reachable within the analysis framework presented in this paper. When $\W$ follows a non-Gaussian distribution, or when different nonlinearities are combined (e.g., $\cos+\sin$ to get the Gaussian kernel \cite{rahimi2008random}), obtaining the equivalent kernel $\bPhi$ in the large $p,T$ regime would be a key enabler to gain a deeper understanding under these more elaborate settings.

Besides, this paper can be taken as a first step of the random matrix-based understanding of various learning methods using random features, for example the randomly designed deep neural networks \cite{lillicrap2016random}, the nonlinear activation of which being the main difficulty for a thorough analysis. Moreover, along with recent advances in random matrix analysis \cite{ali2016spectral}, the hyperparameter $d_1/d_2$ of utmost importance envisions to be consistently estimated and thus allows for an efficient tuning technique for all nonlinear random feature-based methods.

\section*{Acknowledgments}

We thank the anonymous reviewers for their comments and constructive suggestions. We would like to acknowledge this work is supported by the ANR Project RMT4GRAPH (ANR-14-CE28-0006) and the Project DeepRMT of La Fondation Sup{\'e}lec.

\bibliography{liao}
\bibliographystyle{icml2018}

\appendix
\onecolumn

\begin{center}
  {\Large \textbf{Supplementary Material\\}} \vskip 0.1in \textbf{On the Spectrum of Random Features Maps of High Dimensional Data}
\end{center}
\vskip 0.3in

\section{Proof of Theorem~\ref{theo:asymp-equiv-of-Phi-c}}
\label{sm:proof-main-theo}

To obtain the result presented in Theorem~\ref{theo:asymp-equiv-of-Phi-c}, we begin by recalling the expression of the average kernel matrix $\bPhi$, with 
\[
	\bPhi_{i,j} = \bPhi(\x_i, \x_j) = \E_{\w} \G_{ij}  = \E_{\w} \sigma(\w^\T \x_i) \sigma(\w^\T \x_j).
\]

For $\w\sim\mathcal{N}(\mathbf{0},\mathbf{I}_p)$, one resorts to the integral calculus for standard Gaussian distribution in $\mathbb{R}^p$, which can be further reduced to a double integral as shown in \cite{williams1997computing,louart2018random} and results in the expressions in Table~\ref{tab:Phi-sigma}.

Then, from the discussion in Section~\ref{sec:problem}, we have the following expansions for $\x_i \in \mathcal{C}_a$, $\x_j \in \mathcal{C}_b$, with $i \neq j$,
\begin{align}
	\x_i^\T \x_j &= \underbrace{ \bomega_i^\T \bomega_j }_{O(p^{-1/2})} + \underbrace{ \bmu_a^\T \bmu_b /p + \bmu_a^\T \bomega_j/\sqrt{p} + \bmu_b^\T \bomega_i / \sqrt{p} }_{O(p^{-1})}  \label{eq:proof-inner-product} \\
	\|\x_i\|^2 &= \underbrace{\tau}_{O(1)} + \underbrace{ \tr (\C_a^\circ) /p + \|\bomega_i\|^2 - \tr (\C_a) / p}_{O(p^{-1/2})} + \underbrace{ \| \bmu_a\|^2/p + 2\bmu_a^\T \bomega_i /\sqrt{p} }_{O(p^{-1})} \label{eq:proof-norm} 
\end{align}
which further allows one to \emph{linearize} the nonlinear function of $\x_i,\x_j$ in Table~\ref{tab:Phi-sigma} via a Taylor expansion to obtain an \emph{entry-wise} approximation of the key matrix $\bPhi$. 

Nonetheless, this entry-wise approximation does not ensure a vanishing difference in terms of operator norm in the large $p,T$ regime under consideration. Taking the popular Marčenko–Pastur law \cite{marvcenko1967distribution} for example: consider a random matrix $\W \in \mathbb{R}^{n \times p}$ with i.i.d.\@ standard Gaussian entries. Then, as $n,p \to \infty$ with $\frac{n}{p} \to c \in (0,\infty)$, entry-wisely we have that the entry $(i,j)$ of the matrix $\frac1p \W\W^\T$ converge to $1$ if $i=j$ and $0$ otherwise, meaning that the sample covariance matrix $\frac1p \W\W^\T$ seemingly ``converge to'' an identity matrix (which is indeed the true population covariance). But it is well known \cite{marvcenko1967distribution} that the eigenvalue distribution of $\frac1p \W\W^\T$ converges (almost surely so) to a continuous measure (the popular Marčenko–Pastur distribution) compactly supported on $[(1-\sqrt{c})^2, (1+\sqrt{c})^2]$, which is evidently different from the eigenvalue distribution $\delta_{x=1}$ of $\mathbf{I}_n$. 

As a consequence, a more careful control of the entry-wise expansion of $\bPhi$ must be performed to ensure a vanishing expansion error in terms of operator norm. To this end, we follow the previous work of \cite{el2010spectrum,couillet2016kernel} and consider the full matrix contribution.

In the following we proceed the aforementioned manipulations on the $\ReLU$ function as an example, derivations of other functions follow the same procedure and are thus omitted. 

\begin{proof}[Proof of $\sigma(t) = \ReLU(t)$]
We start with the computation of $\bPhi(\ba,\bb)$. For $\sigma(t) = \ReLU(t)$, with the classical Gram-Schmidt process we obtain
\begin{align*}
	\bPhi(\ba,\bb) &= \E_\w \sigma(\w^\T \ba) \sigma(\w^\T \bb) = (2\pi)^{-\frac{p}2} \int_{\mathbb{R}^p} \sigma(\w^\T \ba) \sigma(\w^\T \bb) e^{-\frac12 \|\w\|^2} d\,\w \\
	&= \frac1{2\pi} \int_{\mathbb{R}} \int_{\mathbb{R}} \sigma(\tilde w_1 \tilde a_1) \sigma(\tilde w_1 \tilde b_1 + \tilde w_2 \tilde b_2) e^{-\frac12 (\tilde w_1^2 + \tilde w_2^2) } d\,\tilde w_1 d\,\tilde w_2 \\
	&= \frac1{2\pi} \int_{\mathbb{R}^2} \sigma(\tilde \w^\T \tilde \ba) \sigma(\tilde \w^\T \tilde \bb) e^{-\frac12 \|\tilde \w\|^2} d\,\tilde \w \\
	&= \frac1{2\pi} \int_{ \min(\tilde \w^\T \tilde \ba, \tilde \w^\T \tilde \bb) \ge 0} \tilde \w^\T \tilde \ba \cdot \tilde \w^\T \tilde \bb \cdot e^{-\frac12 \|\tilde \w\|^2} d\,\tilde \w
\end{align*}
where $\tilde a_1 = \|\ba\|$, $\tilde b_1 = \frac{\ba^\T \bb}{\|\ba\|}$, $\tilde b_2 = \|\bb\| \sqrt{ 1 - \frac{(\ba^\T \bb)^2}{ \|\ba\|^2 \|\bb\|^2 } }$ and we denote $\tilde \w = [\tilde w_1, \tilde w_2]^\T$, $\tilde \ba = [\tilde a_1,0]^\T$ and $\tilde \bb = [\tilde b_1,\tilde b_2]^\T$.

With a simple geometric representation we observe 
\[
	\{ \tilde \w~|~\min(\tilde \w^\T \tilde \ba, \tilde \w^\T \tilde \bb) \ge 0 \} = \left\{ r \cos(\theta) + r \sin (\theta)~|~r\ge0,~\theta \in \left[ \theta_0 - \frac{\pi}2, \frac{\pi}2 \right] \right\}
\]
with $\theta_0 \equiv \arccos \left( \frac{ \tilde b_1 }{ \|\bb\| } \right) = \frac{\pi}2 - \arcsin \left( \frac{ \tilde b_1 }{ \|\bb\| } \right)$. Therefore with a polar coordinate change of variable we deduce, for $\sigma(t) = \ReLU(t)$ that
\begin{align*}
	\bPhi(\ba,\bb) &= \tilde a_1 \frac1{2\pi} \int_{\theta_0 - \frac{\pi}2}^{\frac{\pi}2} \cos(\theta) \left( \tilde b_1 \cos(\theta) + \tilde b_2 \sin(\theta) \right) d\,\theta \ \int_{\mathbb{R}^+} r^3 e^{-\frac12 r^2} d\,r \\
	&=\frac1{2\pi} \|\ba\| \|\bb\| \left( \sqrt{ 1-\angle(\ba,\bb)^2 } + \angle(\ba,\bb) \arccos\left( - \angle(\ba,\bb) \right) \right)
\end{align*}
with $\angle(\ba,\bb) \equiv \frac{\ba^\T \bb}{ \|\ba\| \|\bb\| }$ as Table~\ref{tab:Phi-sigma}.

For a second step, with the expressions in \eqref{eq:proof-inner-product} and \eqref{eq:proof-norm} we perform a Taylor expansion to get
\[
	\angle(\x_i,\x_j) = \underbrace{ \frac1{\tau} \bomega_i^\T \bomega_j}_{O(p^{-1/2})} + \underbrace{ \frac1{\tau} \left( \frac1p \bmu_a^\T \bmu_b + \frac1{\sqrt{p}} \bmu_a^\T \bomega_j + \frac1{\sqrt{p}} \bmu_b^\T \bomega_i \right) - \frac{\bomega_i^\T \bomega_j}{2\tau^2} \left(  \frac1p \tr \C_a^\circ + \bphi_i + \frac1p \tr \C_b^\circ + \bphi_j \right) }_{O(p^{-1})} + O(p^{-\frac32})
\]
where we recall $\bphi_i = \| \bomega_i\|^2 - \E \|\bomega_i\|^2 = \| \bomega_i \|^2 - \tr(\C_a)/p $ that is of order $O(p^{-1/2})$. Note that the third term $\bomega_i^\T \bomega_j\left( \tr \C_a^\circ/p + \bphi_i + \tr \C_b^\circ/p + \bphi_j \right)$, being of order $O(p^{-1})$, gives rise to a matrix of vanishing operator norm \cite{couillet2016kernel}, we thus conclude by stating that
\[
	\angle(\x_i,\x_j) = \underbrace{ \frac1{\tau} \bomega_i^\T \bomega_j}_{O(p^{-1/2})} + \underbrace{ \frac1{\tau} \left( \frac1p \bmu_a^\T \bmu_b + \frac1{\sqrt{p}} \bmu_a^\T \bomega_j + \frac1{\sqrt{p}} \bmu_b^\T \bomega_i \right) }_{O(p^{-1})} + O(p^{-\frac32}).
\]

Since $\angle(\x_i,\x_j) = O(p^{-\frac12})$ we sequentially Taylor-expand $\sqrt{1-\angle(\x_i,\x_j)^2}$ and $\arccos(-\angle(\x_i,\x_j))$ to the order of $O(p^{-\frac32})$ as
\begin{align*}
	\sqrt{1-\angle(\x_i,\x_j)^2} = 1 - \frac12 \angle(\x_i,\x_j)^2 + O(p^{-\frac32}) \\
	\arccos(-\angle(\x_i,\x_j)) = \frac{\pi}2 + \angle(\x_i,\x_j) +O(p^{-\frac32}).
\end{align*}

As such, we conclude for $\sigma(t) = \ReLU(t)$ that
\begin{align*}
	\bPhi(\x_i,\x_j) &= \frac1{2\pi} \|\x_i\| \|\x_j\| \left( \sqrt{ 1-\angle(\x_i,\x_j)^2 } + \angle(\x_i,\x_j) \arccos\left( - \angle(\x_i,\x_j) \right) \right) \\
	&= \frac1{2\pi} \|\x_i\| \|\x_j\| \left( 1 + \frac{\pi}2 \angle(\x_i,\x_j) + \frac12 \angle(\x_i,\x_j)^2 \right) + O(p^{-\frac32}) \\
	&= \frac14 \x_i^\T \x_j + \frac1{2\pi} \|\x_i\| \|\x_j\| + \frac1{4\pi} \x_i^\T \x_j \angle(\x_i,\x_j) + O(p^{-\frac32}).
\end{align*}

Consequently we get the generic form for all functions $\sigma(\cdot)$ listed in Table~\ref{tab:Phi-sigma} is given by
\begin{align*}
    &\bPhi_{i,j} \equiv \bPhi(\x_i,\x_j) = \underbrace{ c_0 }_{O(1)} + \underbrace{ c_1 \left( \t_a/\sqrt{p}+ \bphi_i + \t_b/\sqrt{p} + \bphi_j \right) + c_2 \bomega_i^\T \bomega_j }_{O(p^{-1/2})}\\
    &+ \underbrace{ c_3 \left( \t_a/\sqrt{p}+ \bphi_i \right) \left(\t_b/\sqrt{p} + \bphi_j \right) + c_4 \left( ( \t_a/\sqrt{p}+ \bphi_i )^2 + (\t_b/\sqrt{p} + \bphi_j )^2 \right) + c_5 (\bomega_i^\T \bomega_j)^2 }_{O(p^{-1})} \\
    &+ \underbrace{ c_7 \left( (\| \bmu_a\|^2 + \| \bmu_b\|^2) /p + 2(\bmu_a^\T \bomega_i + \bmu_b^\T \bomega_j )/\sqrt{p} \right) + c_8 \left( \bmu_a^\T \bmu_b /p + \bmu_a^\T \bomega_j/\sqrt{p} + \bmu_b^\T \bomega_i / \sqrt{p} \right) }_{O(p^{-1})} + O(p^{-\frac32}) 
\end{align*}
where we recall the definition $\t \equiv \{\tr \C_a^\circ/\sqrt{p}\}_{a=1}^K$. In particular, we have for the $\ReLU$ nonlinearity $c_0 = \frac{\tau}{2\pi}$, $c_1 = c_7 = \frac1{4\pi}$, $c_2 = c_8 = \frac14$, $c_3 = \frac1{8\pi\tau}$, $c_4 = -\frac1{16\pi\tau}$, $c_5 = \frac1{4\pi\tau}$ and $c_6 = 0$.

We then observe that, for all functions $\sigma(\cdot)$ listed in Table~\ref{tab:Phi-sigma}, we have $c_7 = c_1$ and $c_8 = c_2$. Besides, using the fact that
\[
	\left( \bomega_i^\T \bomega_j \right)^2 = \tr(\C_a \C_b)/p^2 + O(p^{-\frac32})
\] 
and considering also the diagonal terms (with $i=j$) by adding the term $c_9 \mathbf{I}_T$, we finally get
\begin{align*}
    \bPhi &= c_0 \mathbf{1}_T \mathbf{1}_T^\T + c_1 \left( \bphi \mathbf{1}_T^\T + \mathbf{1}_T \bphi^\T + \left\{ \frac{\t_a \mathbf{1}_{T_a}}{\sqrt{p}} \right\}_{a=1}^K \mathbf{1}_T^\T + \mathbf{1}_T \left\{ \frac{\t_b \mathbf{1}_{T_b}^\T }{\sqrt{p}} \right\}_{b=1}^K \right) + c_2 \bOmega^\T \bOmega \\
    & + c_3 \left( \bphi \bphi^\T + \bphi \left\{ \frac{\t_b \mathbf{1}_{T_b}^\T }{\sqrt{p}} \right\}_{b=1}^K + \left\{ \frac{\t_a \mathbf{1}_{T_a} }{\sqrt{p}} \right\}_{a=1}^K \bphi^\T + \left\{ \t_a \t_b \frac{\mathbf{1}_{T_a} \mathbf{1}_{T_b}^\T }{p} \right\}_{a,b=1}^K \right)\\
    &+ c_4 \left( (\bphi^2) \mathbf{1}_T^\T + \mathbf{1}_T (\bphi^2)^\T + 2 \left( \mathcal{D}\left\{ \t_a \mathbf{1}_{T_a} \right\}_{a=1}^K \bphi \frac{\mathbf{1}_T^\T}{\sqrt{p}} \right) + 2\left( \frac{\mathbf{1}_T}{\sqrt{p}} \bphi^\T \mathcal{D}\left\{ \t_b \mathbf{1}_{T_b} \right\}_{b=1}^K \right) \right. \\
    &\left. + \left\{ \t_a^2 \frac{\mathbf{1}_{T_a}}{p
    } \right\}_{a=1}^K \mathbf{1}_T^\T +  \mathbf{1}_T \left\{ \t_b^2 \frac{\mathbf{1}_{T_b}^\T }{p
    } \right\}_{b=1}^K \right) + c_5 \left\{ \tr( \C_a \C_b ) \frac{\mathbf{1}_{T_a} \mathbf{1}_{T_b}^\T }{p^2} \right\}_{a,b=1}^K \\
    & + c_1 \left( \left\{ \| \bmu_a\|^2 \frac{\mathbf{1}_{T_a}}{p} \right\}_{a=1}^K \mathbf{1}_T^\T + \mathbf{1}_T \left\{ \| \bmu_b\|^2 \frac{ \mathbf{1}_{T_b}^\T }{p} \right\}_{b=1}^K + \frac2{\sqrt{p}} \left\{ \bOmega_a^\T  \bmu_a \mathbf{1}_{T_a}^\T \right\}_{a=1}^K + \frac2{\sqrt{p}} \left\{ \mathbf{1}_{T_b} \bmu_b^\T \bOmega_b \right\}_{b=1}^K \right)\\
    & + c_2 \left( \frac1p \left\{ \mathbf{1}_{T_a}\bmu_a^\T \bmu_b \mathbf{1}_{T_b}^\T \right\}_{a,b=1}^K + \frac1{\sqrt{p}} \left\{ \bOmega_a^\T  \bmu_b \mathbf{1}_{T_b}^\T + \mathbf{1}_{T_a} \bmu_a ^\T \bOmega_b \right\}_{a,b=1}^K \right) + c_9 \mathbf{I}_T + O_{\|\cdot\|}(p^{-\frac12})
\end{align*}
where we denote $\bphi^2 \equiv [\bphi_1^2, \ldots, \bphi_T^2]^\T$ and $O_{\|\cdot\|}(p^{-\frac12})$ a matrix of operator norm of order $O(p^{-1/2})$ as $p \to \infty$.

Recalling that for $\P \equiv \mathbf{I}_T - \frac1T \mathbf{1}_T \mathbf{1}_T^\T$, we have $\P \mathbf{1}_T = \mathbf{1}_T \P = \mathbf{0}$ and therefore
\begin{align*}
  \bPhi_c &\equiv \P \bPhi\P = c_2 \P \bOmega^\T \bOmega \P + c_3 \P \left( \bphi \bphi^\T + \bphi \left\{ \frac{\t_b \mathbf{1}_{T_b}^\T }{\sqrt{p}} \right\}_{b=1}^K + \left\{ \frac{\t_a \mathbf{1}_{T_a} }{\sqrt{p}} \right\}_{a=1}^K \bphi^\T + \left\{ \t_a \t_b \frac{ \mathbf{1}_{T_a} \mathbf{1}_{T_b}^\T }{p} \right\}_{a,b=1}^K \right) \P\\
  &+ c_5 \P \left\{ \tr( \C_a \C_b ) \frac{ \mathbf{1}_{T_a} \mathbf{1}_{T_b}^\T }{p^2} \right\}_{a,b=1}^K \P + c_2 \P \left( \frac1p \left\{ \mathbf{1}_{T_a}\bmu_a^\T \bmu_b \mathbf{1}_{T_b}^\T \right\}_{a,b=1}^K + \frac1{\sqrt{p}} \left\{ \bOmega_a^\T  \bmu_b \mathbf{1}_{T_b}^\T + \mathbf{1}_{T_a} \bmu_a ^\T \bOmega_b \right\}_{a,b=1}^K \right) \P \\
  & + c_7 \P + O_{\|\cdot\|}(p^{-1/2}) \equiv \P \tilde \bPhi \P + O_{\|\cdot\|}(p^{-1/2}) .
\end{align*}

We further observe that, for all functions $\sigma(\cdot)$ listed in Table~\ref{tab:Phi-sigma} we have $c_5 = 2 c_3$ and let $d_0 = c_7,\ d_1 = c_2,\ d_2 = c_3 = c_5/2$ we obtain the expression of $\tilde \bPhi$ in Theorem~\ref{theo:asymp-equiv-of-Phi-c}, which concludes the proof.
\end{proof}

\end{document}